# Image Restoration from Patch-based Compressed Sensing Measurement

Guangtao Nie[1], Ying Fu[1], Yinqiang Zheng[2], Hua Huang[1]
[1]Beijing Institute of Technology, [2]National Institute of Informatics
{lightbillow,fuying,huahuang}@bit.edu.cn, yqzheng@nii.ac.jp

## Abstract

*A series of methods have been proposed to reconstruct an image from compressively sensed random measurement, but most of them have high time complexity and are inappropriate for patch-based compressed sensing capture, because of their serious blocky artifacts in the restoration results. In this paper, we present a non-iterative image reconstruction method from patch-based compressively sensed random measurement. Our method features two cascaded networks based on residual convolution neural network to learn the end-to-end full image restoration, which is capable of reconstructing image patches and removing the blocky effect with low time cost. Experimental results on synthetic and real data show that our method outperforms state-of-the-art compressive sensing (CS) reconstruction methods with patch-based CS measurement. To demonstrate the effectiveness of our method in more general setting, we apply the de-block process in our method to JPEG compression artifacts removal and achieve outstanding performance as well.*

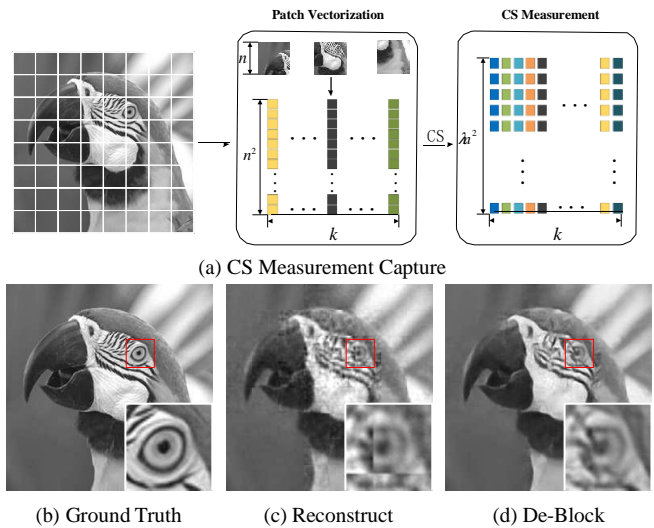

Figure 1: The capturing process of CS measurement (a) and the representative reconstruction (c) and de-block (d) performance of our method. The MR here is 0.10.

## 1. Introduction

Compressive sensing (CS) [8, 1] shows that a sparse signal can be effectively restored with a much lower sampling rate compared with that required by the traditional Shannon sampling theory [12]. Since natural images are intrinsically sparse in some domains [24], they can be effectively reconstructed from CS measurement. Most of existing image reconstruction methods with CS measurement are iterative, which typically need dozens or even hundreds of iterations in the reconstruction process. This limits their application in real-time reconstruction tasks. Besides, these methods often require high measurement rate (MR).

The great success of deep learning in various low-level and high-level computer vision tasks, such as image classification [15, 29, 31, 11], object detection [26, 25], action recognition [28], image segmentation [20], image super-resolution [5, 14, 6] and image deblurring [27], has also been partially generalized into CS image reconstruction, as for example evidenced in the two state-of-the-art non-iterative methods [23, 16]. Compared with their iterative counterparts, these methods have lower time complexity in the online reconstruction process, and also work reasonably well under low MR. However, when using patch-based CS measurement, the reconstruction results of current iterative and non-iterative methods usually suffer from serious blocky artifacts, even though a separate de-block module can be used to alleviate them.

In this paper, we present a non-iterative image restoration method based on residual convolution neural network (CNN) from patch-based CS random measurement. Our method incorporates the patch reconstruction and de-block process into an end-to-end model, which can directly restore the full image without blocky artifacts from patch-based CS input. In order to get a proper tradeoff between restoration quality and time complexity, we design the corresponding network depth for different MR, and achieve outstanding performance in patch reconstruction and blocky artifacts removal, as shown in Figure 1. To further show the effectiveness of the proposed method, we also apply the deblock process in our method into JPEG artifacts removal, which is shown to be superior over the state-of-the-art methods customized for this task.

In summary, our main contributions are that we

1. Present a non-iterative end-to-end full image restoration pipeline for patch-based CS measurement under tight time complexity restriction;
2. Design a deep network structure based on residual CNN, which performs well for both image patch reconstruction and blocky artifacts removal;
3. Demonstrate the effectiveness of our method on synthetic and real patch-based CS data, and its extensibility into JPEG compression artifacts removal.

## 2. Related Work

In the following, we will review most relevant studies on traditional and deep learning based methods for CS reconstruction, as well as blocky artifacts removal methods.

### 2.1. Traditional Methods for CS Reconstruction

Many methods have been proposed for CS reconstruction. For example, Donoho [8] proposed the CS theory and developed the sparse solver with $l_1$-minimization under the assumption that natural images are sparse in some transform domains. Later, various methods, such as K-SVD [9] and stochastic approximations [21], were proposed to adaptively learn the transform domains.

Recently, more constraints have been used to augment the original sparse model [8] for high quality CS reconstruction. Li *et al.* [18] employed total variation minimization to perform CS reconstruction. Dong *et al.* [7] modeled the CS reconstruction by involving a non-local regularization into the optimization function. Metzler *et al.* [22] incorporated a denoising method into the CS reconstruction to effectively mitigate effects from noise in CS reconstruction process.

The algorithms underlying the aforementioned methods are iterative, thus can hardly meet the real-time requirement. Besides, these methods often require high MR and perform much worse for low MR.

### 2.2. Deep Learning for CS Reconstruction

Nowadays, some non-iterative reconstruction methods have been proposed on the basis of deep learning. Mousavi *et al.* [23] used stacked denoising autoencoder to recover a sparse signal from its CS measurement. To reconstruct an image from this autoencoder, many weights are required in the hidden layer. Kulkarni *et al.* [16] employed CNN for CS reconstruction, which effectively reduced the number of learned parameters. These methods can retain rich semantic content at low measurement rate compared with traditional methods for patch-based CS measurement.

Generally, the CS reconstruction is performed on small patches in the image. Therefore, the reconstruction results using non-overlapping patches usually suffer from obvious blocky artifacts, which require an add-on for artifacts removal. To use overlapped patches might alleviate the blocky artifacts, which inevitably requires a higher MR.

### 2.3. Blocky Artifacts Removal

Foi *et al.* [10] constructed an adaptive local filter by adjusting the filter kernel size to remove block edges and preserve image details. Sun and Cham [30] modeled the natural image as a high order Markov random field and the distortion as Gaussian noise, which were involved into an energy function to reduce block distortions. Li *et al.* [19] presented a structure-texture decomposition method to remove the compression artifacts that were amplified in the image contrast enhancement operation. Dong *et al.* [4] produced a CNN model to reduce the compression artifacts. BM3D [3] is an effective and robust denoising method. The deep learning based CS reconstruction method [16] employed BM3D [3] as a denoiser to remove the blocky artifacts. Considering that the time complexity of BM3D is nontrivial, this add-on in effect undermines the benefit of developing a non-iterative CS reconstruction method. In addition, its effectiveness under low MR will deteriorate.

## 3. Residual CNN based CS Restoration

In this section, we develop an end-to-end full image restoration method based on residual CNN, which can directly reconstruct the full image and remove blocky artifacts from patch-based CS measurement under tight time complexity restriction. The overview of our method is shown in Figure 2.

### 3.1. Residual CNN based Network Module

In contrast to traditional CNN, [11] shows that residual CNN can preserve some information in previous layers. In our task, we attempt to employ this property to recover more image details (e.g., edges). Besides, residual CNN can improve the convergence rate and accelerate the training process. Therefore, we design our own network module on the basis of residual CNN, which is referred to as **ResConv** in the following.

As shown in Figure 3, the first layer of ResConv uses kernel size 11 × 11 and generates 64 feature maps. The second layer uses 1 × 1 kernel size and generates 32 feature maps. The final reconstruct layer uses 7 × 7 kernel size and generates only one feature map, which is the output of this module. All the convolutional layers have the same stride of 1, without pooling operation, so as to guarantee that the final output size keeps unchanged. Nonlinear function *ReLU* is used after each convolutional layer except the output layer.

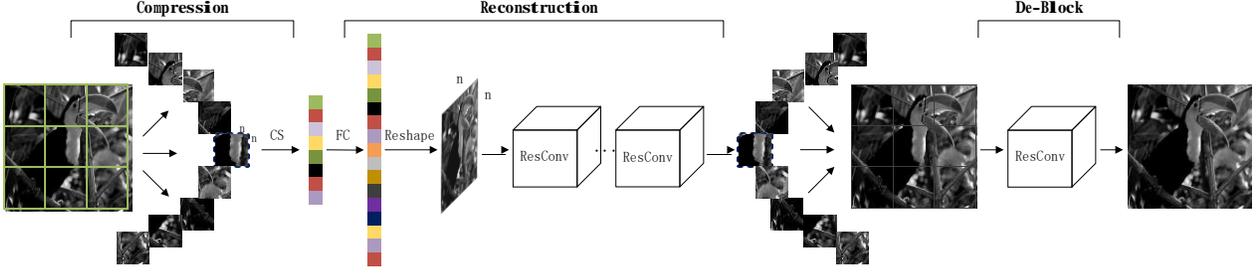

Figure 2: Our proposed restoration pipeline including patch reconstruction and de-blocking.

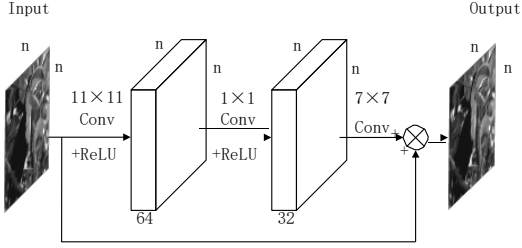

Figure 3: Residual CNN based network module – ResConv.

We can regard the convolutional layer with *ReLU* operation as a nonliner unit, which can be described as

$$\mathbf{Y} = f(\mathbf{X}) = max(0, W * \mathbf{X} + B), \quad (1)$$

where **X** is the input of the convolutional layer and **Y** is the output. *W* is the weight and *B* is the bias of convolutional layer.

### 3.2. Patch Reconstruction from Patch-based Measurement

Let $n \times n$ denote the extracted patch size. So, the number of pixels in one patch is $n^2$. Given a compressive sensing MR of $\lambda$, the length of the sensed vector is $\lambda n^2$.

As illustrated in Figure 2, in the reconstruction process, this vector is first fed into a fully connected (FC) layer, whose output length is $n^2$. This output is reshaped to $n \times n$ and used as the input for ResConv.

We conduct lots of experiments to examine the effect of the number of cascaded ResConv modules, and empirically find that the required depth of the cascaded network is dependent on the measurement rate. In general, one ResConv module already performs very well for high MR (e.g. $\lambda = 0.1$) reconstruction, and to increase the depth further cannot improve the CS reconstruction quality. On the contrary, in the presence of low MR (e.g. $\lambda = 0.01$) input, a cascaded network with multiple ResConv modules can slightly improve the reconstruction performance. Given both the time complexity and reconstruction quality, we thus use two cascaded ResConv modules to reconstruct the patch for low MR.

### 3.3. Blocky Artifacts Removal

When all reconstructed non-overlapping patches are assembled into a full image, the resulting image appears to be blocky. The most relevant study [16] employed an existing denoiser, i.e. BM3D, to remove blocky artifacts. BM3D performs well for high MR reconstruction, but can not effectively remove the blocky artifacts when MR is lower than 0.1. In our method, we attempt to use deep learning to remove the blocky artifacts as well, and construct an end-to-end CS restoration model on the basis of ResConv.

We use one ResConv module only for artifacts removal, since our empirical evaluation shows that one ResConv module performs better than cascaded modules for the de-block process.

This de-block process has three major effects. Firstly, it removes the blocky artifacts as expected; Secondly, it can alleviate the noise originated in reconstruction process; Finally, it can predict the high frequency information of the image and further restore image details. Because all the layers in this network are convolutional layers, there is no restriction on the input size. After training this network, it is able to handle images of any size.

As will be shown in the experiment section, this residual CNN based de-block process outperforms traditional denoisers. So, it can also be used as an add-on for existing patch-based CS reconstruction method to further improve the reconstruction performance on full images.

### 3.4. Training Details

Learning the end-to-end mapping function $f$ requires to estimate the network parameters $\{W, B\}$ firstly. This can be achieved by minimizing the loss between the reconstructed image $f(\mathbf{Y}; W, B)$ and the corresponding ground truth image **X**. The Mean Squared Error (MSE) is employed as the loss function,

$$\text{Loss} = \frac{1}{k} \sum_{i=1}^{k} \| f(\mathbf{Y}_i; W, B) - \mathbf{X}_i \|^2, \quad (2)$$

where $\mathbf{Y}_i$ is the $i^{th}$ input and $\mathbf{X}_i$ is the $i^{th}$ corresponding ground truth. $k$ is the number of training samples. The loss is minimized with the stochastic gradient descent (SGD) method [17]. The input and output of the network are single channel images.

Table 1: Evaluate on PSNR of test images and running time in seconds of 256 × 256 images for different model parameter selection.

| Model | MR=0.25 | | MR=0.10 | | MR=0.04 | | MR=0.01 | |
|---|---|---|---|---|---|---|---|---|
| | PSNR | Time | PSNR | Time | PSNR | Time | PSNR | Time |
| ReconNet | 25.5459 | 0.008 | 23.1522 | 0.008 | 20.9234 | 0.007 | **17.9023** | 0.008 |
| Half-ReconNet | 26.6286 | 0.005 | 23.5820 | 0.005 | 21.0520 | 0.005 | 17.7952 | 0.005 |
| FC-2-ResConv | 26.8760 | 0.008 | 23.5960 | 0.008 | 20.9976 | 0.007 | 17.8929 | 0.008 |
| FC-1-ResConv | **27.2172** | 0.005 | **23.6113** | 0.005 | **21.2171** | 0.005 | 17.7912 | 0.005 |

Table 2: Comparison of BM3D and our ResConv method for de-blocking. We evaluate PSNR using four different methods for patch reconstruction.

| Algorithm | MR=0.25 | | MR=0.10 | | MR=0.04 | | MR=0.01 | |
|---|---|---|---|---|---|---|---|---|
| | BM3D | ResConv | BM3D | ResConv | BM3D | ResConv | BM3D | ResConv |
| TVAL3 | 27.6086 | **28.8711** | 23.2905 | **23.8735** | 19.6797 | **20.3906** | 15.6706 | **17.1215** |
| D-AMP | 27.4477 | **28.2917** | 20.2199 | **21.5525** | 14.2572 | **17.8176** | 5.3887 | **13.2846** |
| ReconNet | 25.9285 | **26.6449** | 23.5603 | **24.0359** | 21.1909 | **21.4255** | 17.9993 | **18.2832** |
| Ours | 27.3472 | **28.5301** | 23.9740 | **24.7082** | 21.4180 | **21.7270** | 17.9848 | **18.3082** |

The extracted patch size $n = 32$, and four different MRs are used, i.e. $\lambda = 0.25, 0.10, 0.04$ and $0.01$. Thus, the number of measurements is 256, 102, 40 and 10 for patches under different MRs, respectively.

Each convolution layer's weights are initialized by random sampling from a Gaussian distribution with zero mean and fixed standard deviation. Similar with [16], for the patch-based CS reconstruction, the initialized standard deviation for the fully connected layer is 0.01 and 0.1 for other convolutional layers. The learning rate is different for each layer in the full network. [5] have found that the last layer with smaller learning rate is important for the network to converge. Therefore, we set the learning rate $10^{-5}$ for the first two convolutional layers and $10^{-6}$ for the last layer. For the de-block network, the standard deviation for all the convolutional layers is 0.001. The learning rate is $10^{-3}$ for the first two layers and $10^{-4}$ for the last layer.

The momentum for both patch-based CS reconstruction and de-block networks is 0.9, and the biases are initialized to be zero. All the networks have been trained with the deep learning tools Caffe [13] on the NVIDIA Titan X GPU.

## 4. Experimental Results

In this section, we will firstly introduce the generation of our training dataset and discuss the setting of model parameters. Then, qualitative evaluation on synthetic data is shown. To demonstrate the effectiveness of our approach, we also perform the proposed method on the real capture data and extend the block removal process to the task of compression artifacts removal in JPEG images.

### 4.1. Training Dataset

We use the same set of 91 images as in [5, 16] to generate our training set. The Set 5 [2] constitutes our validation set, which is used to evaluate the performance of our model during the training process. In this paper, we use the luminance channel of each image to construct the training dataset.

Our restoration pipeline includes CS reconstruction from patch-based measurement and blocky artifact removal. Thus, different training datasets are generated for these two parts, respectively.

#### 4.1.1 Dataset for Patch Reconstruction

To formulate the training and validation datesets, we uniformly extract patches with the size of 32 × 32. The stride of extraction is 14 for training and 21 for validation. Thus, the training dataset has 22144 patches and the validation set includes 1112 patches. These patches constitute the ground truth. Then, we conduct a measurement matrix $\varphi$. The size of $\varphi$ is $\lambda n^2 \times n^2$, where $n = 32$ and $\lambda$ is the measurement rate. We use four MR = 0.25, 0.10, 0.04 and 0.01 as mentioned above. $\varphi$ is generated from a random Gaussian matrix with appropriate size, and its rows are orthonormalized.

The input of the patch-based CS reconstruction network reads

$$\mathbf{y} = \varphi \mathbf{x}_{vec}, \qquad (3)$$

where $\mathbf{x}_{vec}$ is the vectorized version of the input image patch $\mathbf{x}$, the length of $\mathbf{x}_{vec}$ is $n^2$, and the training set is labeled as $(\mathbf{y}, \mathbf{x})$.

Four different training datasets are produced for different MRs, and they are used to train our patch-based CS reconstruction network. We thus obtain four models corresponding to four different rates.

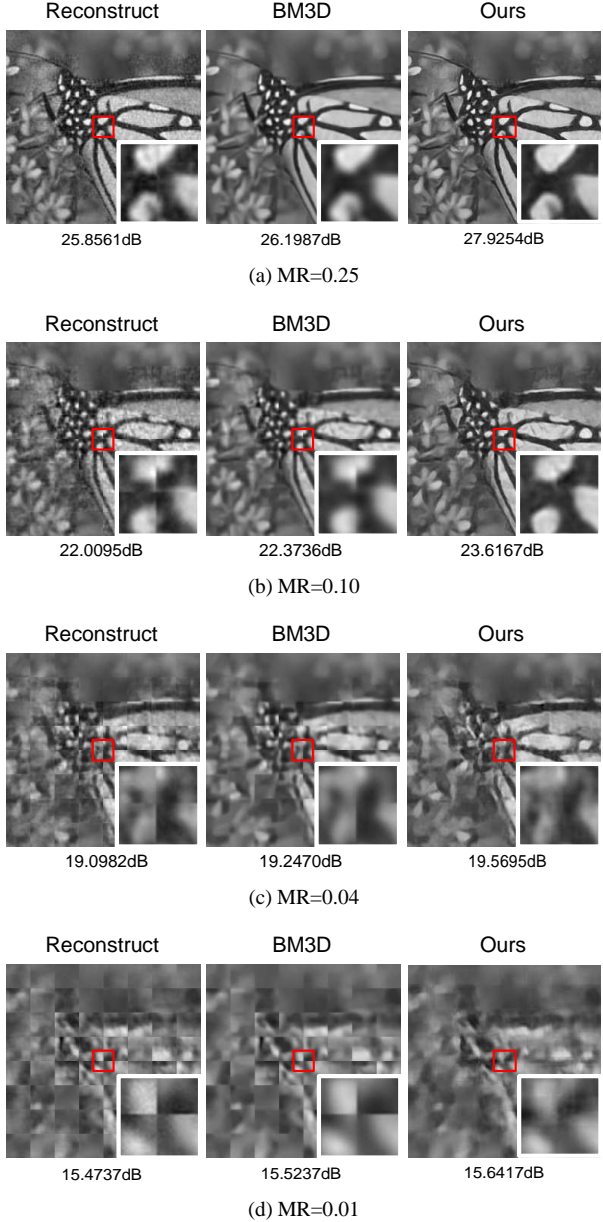

Figure 4: Comparison of BM3D and our de-block method when used to remove the blocky effects at different measurement rates.

### 4.1.2 Dataset for Blocky Artifact Removal

To generate the blocky artifacts removal dataset, we extract non-overlapping patches of size 32 × 32 from the original image, compress each patch with CS measurement, and reconstruct them with our patch-based CS reconstruction network mentioned above. The reconstructed non-overlapping patches are assembled into a full image, which contains serious blocky artifacts. By repeating this operation on all images in the training set, we obtain 91 blocky images.

Then, we extract patches $\mathbf{x}_{block}$ in the same way as preparing the reconstruction dataset. Since the patch size $n = 32$ is not divisible for stride 14 and 21, almost all extracted patches contain blocky artifacts. Besides, the blocky artifacts location would be different for extracted patches, which ensures the diversity of datasets. These overlapped patches make up the inputs of our de-block network. The ground truth is the same as the reconstruct net dataset $\mathbf{x}$. We label the de-block training set as $(\mathbf{x}_{block}, \mathbf{x})$.

Note that, according to the MR, the dataset for de-block network should also be different. We therefore generate different datasets for their corresponding measurement rates, separately.

### 4.2. Evaluation on Patch-based CS Reconstruction

For the simulated data in all our experiments, we evaluate the proposed method on the same test images as in [16], which consists of 11 grayscale images, with 9 images of size 256 × 256 and 2 images of size 512 × 512. In the following experiments, we compute the average PSNR value for the total 11 images and the average running time for the 9 images of size 256 × 256.

We first examine the effect of different network structures and depths for patch-based CS reconstruction under different MRs. ReconNet is the same network as in [16]. Half-ReconNet is a compact version of ReconNet by removing the last three convolution layers. FC-1-ResConv denotes one ResConv module with a fully connected layer in precedence. FC-2-ResConv denotes two cascaded ResConv modules with a fully connected layer. The results of the patch-based CS reconstruction for different network structures are shown in Table 1. When MR=0.01, the best network is ReconNet and results from FC-2-ResConv are slightly better than those from FC-1-ResConv. This observation implies that a deeper network performs better when MR is very low. In all other cases, FC-1-ResConv network performs best, which indicates that in general, only one ResConv module in the network is enough to reconstruct the patch under CS measurement.

As for time complexity, the running speed of ResConv is similar to that of ReconNet with the same depth. FC-1-ResConv and Half-ReconNet are almost twice as fast as deeper models. Given the tradeoff between reconstruction performance and time complexity, we select the FC-1-ResConv model in general and the FC-2-ResConv model for a very low MR (e.g. $\lambda = 0.01$).

### 4.3. Evaluation on Blocky Artifacts Removal

In [16], BM3D follows ReconNet to remove blocky artifacts. In Table 2, we mainly compare our de-block network with BM3D for blocky artifacts removal. Three state-of-the-art CS reconstruction methods under patch-based measurement are used here, including TVAL3 [18], D-

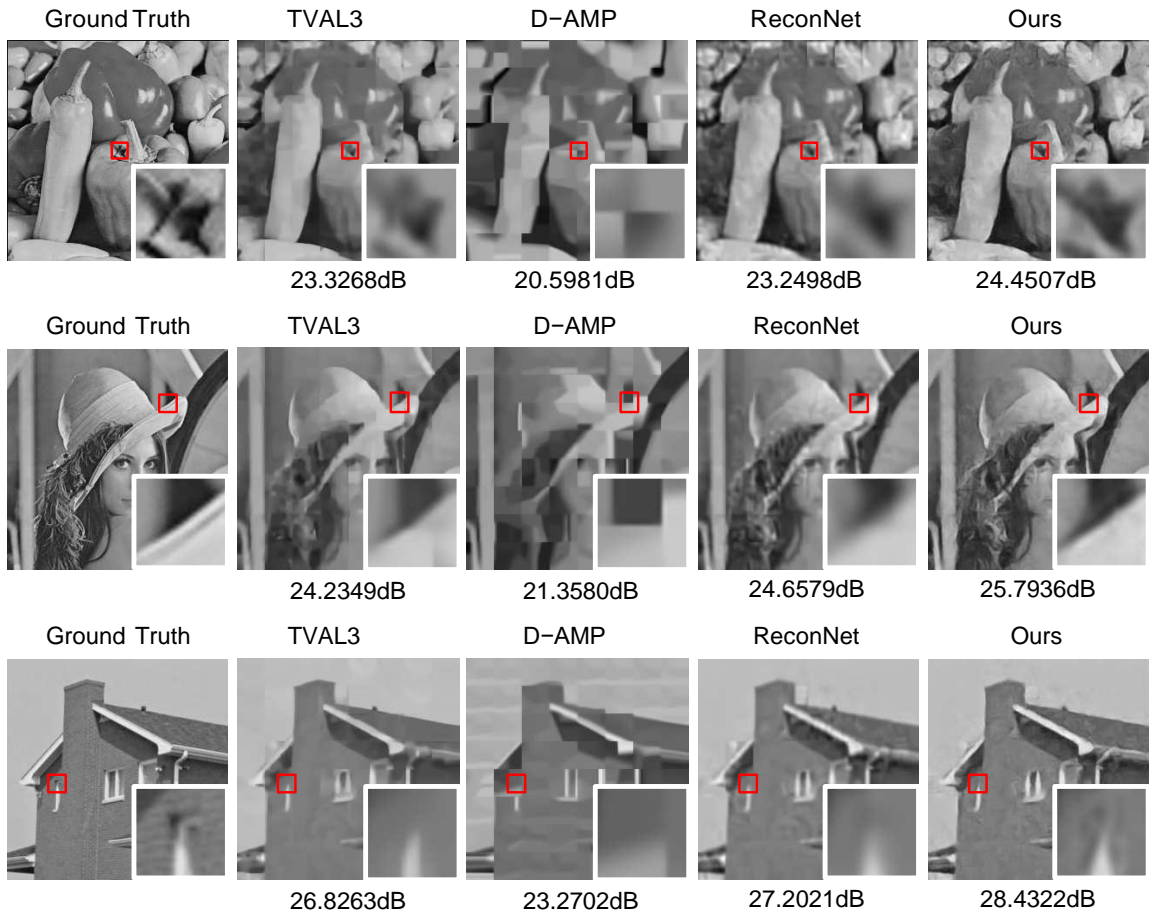

Figure 5: Final restoration samples of different methods with block removal at MR=0.10. BM3D is used in competing methods for block removal, while the ResConv model is used for our block removal.

AMP [22], and ReconNet [16]. We can see that our de-block network outperforms BM3D for all the three CS reconstruction methods.

For illustration, we show the de-block results of BM3D and our de-block network for the blocky images produced by our patch-based CS reconstruction method under different MRs in Figure 4. We can see that BM3D performs well when the MR is high, but its de-block results still suffer from blocky artifact for low MRs. Our de-block network performs better for all MRs and provides more shape details in the final results.

### 4.4. Comparison of End-to-End Full Restoration

Here, we compare our approach with three state-of-the-art methods, including two traditional CS reconstruction methods ( TVAL3 [18] and D-AMP [22] ) and a deep learning based method ReconNet [16], in term of end-to-end full restoration performance. All competing methods employs BM3D [3] for the de-block process, while our method uses the aforementioned network. All methods are qualitatively evaluated by measuring the PSNR and the running time in seconds. We use the same compressively sensed random measurements for all compared methods. In Table 3, we show the evaluation results for 5 images and provide the mean PSNR for all images in the testing dataset.

We can see that deep learning based methods often perform better than traditional CS reconstruction methods under patch-based measurement, except that MR is 0.25. In terms of the two learning based methods, our method outperforms ReconNet in most cases, but is slightly worse than ReconNet when MR is 0.01. However, after the de-block process, we can see our algorithm performs best among all the compared methods for all MRs. Three samples at MR = 0.10 are shown in Figure 5. Compared with other methods, our method produces more details in the final generated pictures.

Table 4 shows the average time complexity for 256×256 image of all compared methods. We can see that our method has shorter running time than the three competing methods. These evaluations have demonstrated the effectiveness and efficiency of our proposed method.

Table 3: Comparison between our method and existing ones using different algorithms at different measurement rates. BM3D is used for previous methods for block removal, while ResConv is used for our block removal. We compute mean PSNR value with all the 11 test images.

| picture | Algorithm | MR=0.25 | | MR=0.10 | | MR=0.04 | | MR=0.01 | |
|---|---|---|---|---|---|---|---|---|---|
| | | Reconstruct | Block Remove | Reconstruct | Block Remove | Reconstruct | Block Remove | Reconstruct | Block Remove |
| 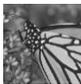 | TVAL3 | **27.7400** | 27.1806 | 20.9922 | 21.4172 | 17.3358 | 17.5422 | 13.6735 | 13.7573 |
| | D-AMP | 26.5705 | 25.9058 | 18.4640 | 18.3990 | 14.0495 | 13.9152 | 6.4607 | 6.4384 |
| | ReconNet | 23.9278 | 24.3815 | 21.4352 | 21.8109 | 18.7001 | 18.8474 | 15.4344 | 15.4877 |
| | Ours | 25.8561 | **27.9254** | **22.0095** | **23.6167** | **19.0982** | **19.5695** | **15.4737** | **15.6417** |
| 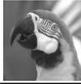 | TVAL3 | 27.0146 | 27.2521 | 23.3383 | 23.7386 | 20.2940 | 20.6357 | 16.1746 | 16.3014 |
| | D-AMP | 26.3903 | 26.1105 | 21.0717 | 21.1997 | 14.9773 | 13.9325 | 5.3321 | 5.3126 |
| | ReconNet | 25.4951 | 25.7676 | 23.3751 | 23.7141 | 21.8940 | 22.1238 | **19.2428** | 19.3654 |
| | Ours | **27.2395** | **28.4585** | **24.0308** | **24.9242** | **22.0241** | **22.3591** | 19.1768 | **19.7481** |
| 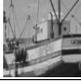 | TVAL3 | 28.6675 | 28.3067 | 23.6453 | 24.0747 | 20.2760 | 20.5544 | 16.0676 | 16.2306 |
| | D-AMP | **29.1559** | 28.3220 | 21.0940 | 21.0898 | 14.8545 | 14.9804 | 5.5725 | 5.5397 |
| | ReconNet | 27.0910 | 27.2909 | 24.3535 | 24.6834 | 22.0232 | 22.3106 | **18.6871** | 18.7933 |
| | Ours | 28.5385 | **29.7123** | **24.7562** | **25.7705** | **22.3830** | **22.8226** | 18.6825 | **19.0747** |
| 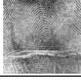 | TVAL3 | 22.7216 | 23.2908 | 18.5989 | 18.9084 | 16.3993 | 16.5071 | 13.5555 | 13.6345 |
| | D-AMP | 24.9449 | 24.2569 | 16.3546 | 16.3073 | 12.9779 | 13.0233 | 4.9674 | 4.8652 |
| | ReconNet | 24.9941 | 24.9610 | 20.7998 | 20.9898 | 17.2718 | 17.3177 | 15.0251 | 15.0428 |
| | Ours | **26.9628** | **27.0305** | **21.3349** | **21.9216** | **17.4230** | **17.6853** | **15.0408** | **15.2393** |
| 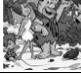 | TVAL3 | **23.8337** | 24.5624 | 18.7738 | 19.2193 | 15.7998 | 15.9804 | 12.3468 | 12.4396 |
| | D-AMP | 24.7688 | 24.3754 | 16.6163 | 16.5357 | 12.3578 | 12.2899 | 4.5302 | 4.5140 |
| | ReconNet | 21.5438 | 22.0976 | 18.7744 | 19.1155 | 16.4678 | 16.6533 | **14.0030** | 14.0642 |
| | Ours | 23.7391 | **25.6027** | **19.3888** | **20.7252** | **16.8458** | **17.2736** | 13.9177 | **14.2895** |
| Mean PSNR | TVAL3 | 27.7025 | 27.6086 | 22.7967 | 23.2906 | 19.4125 | 19.6797 | 15.4811 | 15.6076 |
| | D-AMP | **28.0766** | 27.4478 | 20.1821 | 20.2199 | 14.2305 | 14.2572 | 5.4430 | 5.3887 |
| | ReconNet | 25.5459 | 25.9285 | 23.1522 | 23.5603 | 20.9234 | 21.1909 | **17.9023** | 17.9993 |
| | Ours | 27.2172 | **28.5301** | **23.6113** | **24.7082** | **21.2171** | **21.7270** | 17.8929 | **18.3082** |

Table 4: Time complexity for $256 \times 256$ images using different algorithms at different measurement rates. BM3D is used for previous method for block removing, ResConv model is used for our block removal.

| Algorithm | MR=0.25 | | MR=0.10 | | MR=0.04 | | MR=0.01 | |
|---|---|---|---|---|---|---|---|---|
| | Reconstruct | Block Remove | Reconstruct | Block Remove | Reconstruct | Block Remove | Reconstruct | Block Remove |
| TVAL3 | 3.5812 | 4.1603 | 3.9359 | 4.4877 | 4.4821 | 5.0103 | 5.0879 | 5.6544 |
| D-AMP | 26.5497 | 27.2017 | 33.9030 | 33.9487 | 38.3642 | 38.7258 | 39.4764 | 39.9886 |
| ReconNet | 0.0079 | 0.5492 | 0.0073 | 0.5539 | 0.0076 | 0.5456 | 0.0076 | 0.5534 |
| Ours | **0.0054** | **0.0217** | **0.0049** | **0.0229** | **0.0047** | **0.0221** | **0.0073** | **0.0235** |

### 4.5. Performance on Real Images

We also perform our method on real data captured by a block single pixel camera [16]. This designed capture system consists of two optical arms and a discrete micro-mirror device (DMD) acting as a spatial light modulator, which is used to obtain the CS measurements. 383 patches under CS measurement are captured for each full image and the size of the patch is $33 \times 33$.

Since the block size of these real data is 33×33, we need to retrain our model on patches of this size. We thus generate corresponding training and validation sets by following the protocol mentioned in Section 4.1, while keeping all other parameters unchanged except the patch size.

The deep learning based models are trained on two MRs, i.e. 0.10 and 0.04. To effectively show the comparison results, we also test the real capture data on TVAL3 [18], D-AMP [22] and ReconNet [16]. Again, different from the competing methods using BM3D for artifacts removal, we use our ResConv module to remove the blocky artifacts after patch reconstruction. The restored full images are shown in Figure 6. We can see that our algorithm offers visually better restoration results than the other three methods under different MRs, which verifies the effectiveness of our method for real CS capture data.

### 4.6. Extension on JPEG Images

JPEG is a lossy compression method, which tends to introduce compression artifacts, such as blocky artifacts and ringing effects. The blocky artifacts result from discontinuities at 8×8 borders, while the ringing effects usually appear along strong edges. We have presented a de-block network based on residual CNN in Section 3 for patch-based CS reconstruction. To show the extensibility, we also implement our de-block network to decrease the compression artifacts

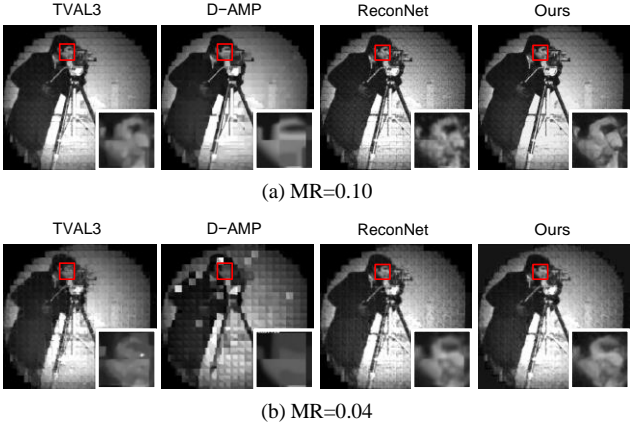

(a) MR=0.10

(b) MR=0.04

Figure 6: Comparison results on real data with different MRs. The restoration results from our method are qualitatively better than those from the competing methods.

Table 5: PSNR comparison of different de-block algorithms for JPEG images at three different quality Q1, Q2 and Q3.

| Quality | Q1 | Q2 | Q3 |
|---|---|---|---|
| JPEG Compressed | 28.0207 | 27.5422 | 24.7479 |
| SA-DCT | 29.4862 | 28.9951 | 26.3670 |
| FoE | 29.1744 | 28.7301 | 26.0072 |
| Li | 28.5778 | 28.2004 | 25.5359 |
| AR-CNN | 29.6337 | 29.1661 | 26.3359 |
| ResConv | **29.7456** | **29.2335** | **26.5167** |

Table 6: Time costs comparison for 256 × 256 images between different de-block algorithms towards JPEG images at three different quality Q1, Q2 and Q3.

| Quality | Q1 | Q2 | Q3 |
|---|---|---|---|
| SA-DCT | 3.0932 | 3.1160 | 3.2651 |
| FoE | 46.9423 | 47.2292 | 46.1712 |
| Li | 2.7990 | 2.8155 | 2.8290 |
| AR-CNN | 0.0211 | 0.0217 | 0.0208 |
| ResConv | **0.0147** | **0.0150** | **0.0147** |

in JPEG images, especially for the blocky artifacts.

Here, we compare our de-block method with four state-of-the-arts de-blocking methods, including FoE [30], SA-DCT [10], Li [19] and AR-CNN [4]. The last one is a deep learning based method.

In the experiment, we use three JPEG quality setting Q1, Q2 and Q3, which are the same as in previous work [10, 30]. We compress our training dateset and get 91 compressed samples. Patches are extracted in the same way as in generating training dataset for de-block process in Section 4.1. The JPEG de-block model is trained on this training dataset. To make fair comparison, we use the same training dataset for AR-CNN [4].

The results on PSNR are shown in Table 5. We can see that our de-block method achieves higher PSNR values than all the competing methods, and has lower time complexity, as shown in Table 6. To visualize the artifacts removal performance, we also show the restored JPEG images for all competing approaches in Figure 7. All these results demonstrate the effectiveness and extensibility of our de-block method.

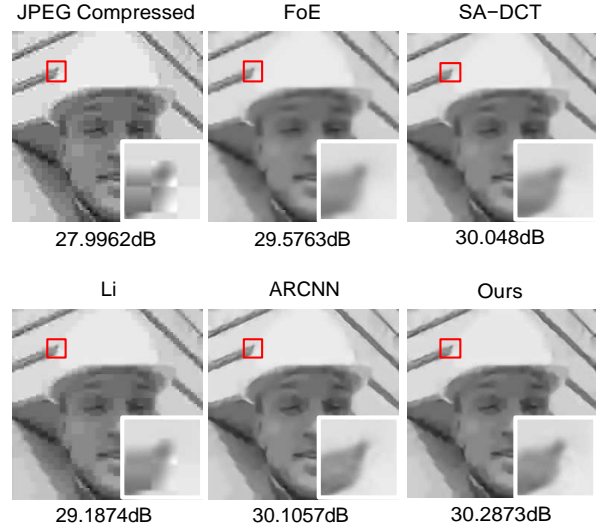

Figure 7: PSNR values of different methods for JPEG artifacts removal. The quality criterion here is Q3.

## 5. Conclusion

In this paper, we have developed a non-iterative image reconstruction method based on residual convolution neural network, which involves the patch reconstruction and de-block process into an integrated end-to-end network. The proposed method can directly reconstruct the full image without blocky artifacts from patch based CS measurement. We have properly designed the network structure and depth for different measurement rates, by trading off restoration quality and time complexity. The effectiveness of our proposed method has been verified by using synthetic and real capture data. We have also extended the de-block process in our proposed method for JPEG compression artifacts reduction, and achieved superior performance compared with the state-of-the-art methods.

Our current network is designed for monochromatic images and it is worth investigating how to extend our method into RGB/mutispectal/hyperspectal image capture and restoration.

## References

[1] R. G. Baraniuk. Compressive sensing [lecture notes]. *IEEE Trans. Signal Processing Magazine*, 24(4):118–121, 2007. 1

[2] M. Bevilacqua, A. Roumy, C. Guillemot, and A. Morel. Low-complexity single-image super-resolution based on nonnegative neighbor embedding. *Proceedings of the British Machine Vision Conference*, 2012. 4

[3] K. Dabov, A. Foi, V. Katkovnik, and K. Egiazarian. Image denoising by sparse 3-d transform-domain collaborative filtering. *IEEE Trans. Image Processing*, 16(8):2080–2095, 2007. 2, 6

[4] C. Dong, Y. Deng, C. L. Chen, and X. Tang. Compression artifacts reduction by a deep convolutional network. In *Proc. of International Conference on Computer Vision (ICCV)*, pages 576–584, 2015. 2, 8

[5] C. Dong, C. C. Loy, K. He, and X. Tang. Image super-resolution using deep convolutional networks. *IEEE Trans. Pattern Analysis and Machine Intelligence (PAMI)*, 38(2):295–307, 2016. 1, 4

[6] C. Dong, C. C. Loy, and X. Tang. Accelerating the super-resolution convolutional neural network. In *Proc. of European Conference on Computer Vision (ECCV)*, pages 391–407. Springer, 2016. 1

[7] W. Dong, G. Shi, X. Li, Y. Ma, and F. Huang. Compressive sensing via nonlocal low-rank regularization. *IEEE Trans. Image Processing*, 23(8):3618–3632, 2014. 2

[8] D. L. Donoho. Compressed sensing. *IEEE Trans. Information Theory*, 52(4):1289–1306, 2006. 1, 2

[9] M. Elad and M. Aharon. Image denoising via sparse and redundant representations over learned dictionaries. *IEEE Trans. Image Processing*, 15(12):3736–3745, Dec. 2006. 2

[10] A. Foi, V. Katkovnik, and K. Egiazarian. Pointwise shape-adaptive dct for high-quality denoising and deblocking of grayscale and color images. *IEEE Trans. Image Processing*, 16(5):1395–1411, 2007. 2, 8

[11] K. He, X. Zhang, S. Ren, and J. Sun. Deep residual learning for image recognition. In *Proc. of IEEE Conference on Computer Vision and Pattern Recognition (CVPR)*, pages 770–778, 2016. 1, 2

[12] A. J. Jerri. The shannon sampling theoremits various extensions and applications: A tutorial review. *Proceedings of the IEEE*, 65(11):1565–1596, 1977. 1

[13] Y. Jia, E. Shelhamer, J. Donahue, S. Karayev, J. Long, R. Girshick, S. Guadarrama, and T. Darrell. Caffe: Convolutional architecture for fast feature embedding. In *Proceedings of ACM international conference on Multimedia*, pages 675–678. ACM, 2014. 4

[14] J. Kim, J. Kwon Lee, and K. Mu Lee. Accurate image super-resolution using very deep convolutional networks. In *Proc. of IEEE Conference on Computer Vision and Pattern Recognition (CVPR)*, pages 1646–1654, 2016. 1

[15] A. Krizhevsky, I. Sutskever, and G. E. Hinton. Imagenet classification with deep convolutional neural networks. In *Proc. of Conference on Neural Information Processing Systems (NIPS)*, pages 1097–1105, 2012. 1

[16] K. Kulkarni, S. Lohit, P. Turaga, R. Kerviche, and A. Ashok. Reconnet: Non-iterative reconstruction of images from compressively sensed measurements. In *Proc. of IEEE Conference on Computer Vision and Pattern Recognition (CVPR)*, June 2016. 1, 2, 3, 4, 5, 6, 7

[17] Y. LeCun, L. Bottou, Y. Bengio, and P. Haffner. Gradient-based learning applied to document recognition. *Proceedings of the IEEE*, 86(11):2278–2324, 1998. 4

[18] C. Li, W. Yin, H. Jiang, and Y. Zhang. An efficient augmented lagrangian method with applications to total variation minimization. *Computational Optimization and Applications*, 56(3):507–530, 2013. 2, 5, 6, 7

[19] Y. Li, F. Guo, R. T. Tan, and M. S. Brown. *A Contrast Enhancement Framework with JPEG Artifacts Suppression*. Proc. of European Conference on Computer Vision (ECCV), 2014. 2, 8

[20] J. Long, E. Shelhamer, and T. Darrell. Fully convolutional networks for semantic segmentation. In *Proc. of IEEE Conference on Computer Vision and Pattern Recognition (CVPR)*, pages 3431–3440, 2015. 1

[21] J. Mairal, F. Bach, J. Ponce, and G. Sapiro. Online dictionary learning for sparse coding. In *Proc. of International Conference on Machine Learning (ICML)*, pages 689–696, 2009. 2

[22] C. A. Metzler, A. Maleki, and R. G. Baraniuk. From denoising to compressed sensing. *IEEE Trans. Information Theory*, 62(9):5117–5144, 2016. 2, 6, 7

[23] A. Mousavi, A. B. Patel, and R. G. Baraniuk. A deep learning approach to structured signal recovery. In *Annual Allerton Conference on Communication, Control, and Computing*, pages 1336–1343. IEEE, 2015. 1, 2

[24] B. A. Olshausen and D. J. Field. Emergence of simple-cell receptive field properties by learning a sparse code for natural images. *Nature*, 381(6583):607–609, June 1996. 1

[25] J. Redmon, S. Divvala, R. Girshick, and A. Farhadi. You only look once: Unified, real-time object detection. In *Proc. of IEEE Conference on Computer Vision and Pattern Recognition (CVPR)*, pages 779–788, 2016. 1

[26] S. Ren, K. He, R. Girshick, and J. Sun. Faster r-cnn: Towards real-time object detection with region proposal networks. In *Proc. of Conference on Neural Information Processing Systems (NIPS)*, pages 91–99, 2015. 1

[27] C. J. Schuler, M. Hirsch, S. Harmeling, and B. Scholkopf. Learning to deblur. *IEEE Trans. Pattern Analysis and Machine Intelligence (PAMI)*, 38(7):1439–1451, 2014. 1

[28] K. Simonyan and A. Zisserman. Two-stream convolutional networks for action recognition in videos. In *Proc. of Conference on Neural Information Processing Systems (NIPS)*, pages 568–576, 2014. 1

[29] K. Simonyan and A. Zisserman. Very deep convolutional networks for large-scale image recognition. *arXiv preprint arXiv:1409.1556*, 2014. 1

[30] D. Sun and W. K. Cham. Postprocessing of low bit-rate block dct coded images based on a fields of experts prior. *IEEE Trans. Image Processing*, 16(11):2743–2751, 2007. 2, 8

[31] C. Szegedy, W. Liu, Y. Jia, P. Sermanet, S. Reed, D. Anguelov, D. Erhan, V. Vanhoucke, and A. Rabinovich. Going deeper with convolutions. In *Proc. of IEEE Conference on Computer Vision and Pattern Recognition (CVPR)*, pages 1–9, 2015. 1